\documentclass[conference]{IEEEtran}
\IEEEoverridecommandlockouts
\usepackage{cite}
\usepackage{amsmath,amssymb,amsfonts}
\usepackage{algorithmic}
\usepackage{float}
\usepackage{graphicx}
\usepackage{textcomp}
\usepackage[table,xcdraw, dvipsnames]{xcolor}
\usepackage{url}
\usepackage{multirow}
\usepackage{hyperref}
\hypersetup{
 colorlinks=True,
}

\usepackage{booktabs}

\begin{document}

\title{End-to-end Sleep Staging with Raw Single Channel EEG using Deep Residual ConvNets\\
\thanks
}

\author{Ahmed Imtiaz Humayun$^{1}$,
        Asif Shahriyar Sushmit$^{1}$,
        Taufiq Hasan$^{1}$, 
        and Mohammed Imamul Hassan Bhuiyan$^{1,2}$

\thanks{$^{1}$mHealth Research Group, Department of Biomedical Engineering, Bangladesh University of Engineering and Technology (BUET), Dhaka - 1205, Bangladesh. Email: {\tt\scriptsize taufiq@bme.buet.ac.bd}}%
\thanks{$^{2}$Department of Electrical and Electronics Engineering, Bangladesh University of Engineering and Technology, Dhaka - 1205, Bangladesh Email: {\tt\scriptsize imamul@eee.buet.ac.bd }}}

\maketitle
\begin{abstract}
Humans approximately spend a third of their life sleeping, which makes monitoring sleep an integral part of well-being. In this paper, a 34-layer deep residual ConvNet architecture for end-to-end sleep staging is proposed. The network takes raw single channel electroencephalogram (Fpz-Cz) signal as input and yields hypnogram annotations for each 30s segments as output. Experiments are carried out for two different scoring standards (5 and 6 stage classification) on the expanded PhysioNet Sleep-EDF dataset, which contains multi-source data from hospital and household polysomnography setups. The performance of the proposed network is compared with that of the state-of-the-art algorithms in patient independent validation tasks.  
The experimental results demonstrate 
the superiority of the proposed network 
compared to the best existing method, providing a relative improvement in epoch-wise average accuracy of 6.8\% and 6.3\% on the household data and multi-source data, respectively. Codes are made publicly available on \mdseries \texttt{\href{https://github.com/mHealthBuet/ASSC}{Github}}.
\end{abstract}


\section{Introduction}
Sleep disorders are becoming a global health problem and more prominent in clinical practice as the global population and life expectancy increases. 50-70 million adults have a sleep disorder in the United States according to the American Sleep Association. Sleep problems were marked as ``An Emerging Global Epidemic" following a study by the World Health Organization (WHO) conducted with 40,000 adults from developing countries across Africa and Asia \cite{stranges2012sleep}. 
Sleep plays an essential role in our health and well-being. It affects the regulation of glucose metabolism and loss of sleep could develop a resistance towards insulin, which may result in Type 2 diabetes \cite{spiegel2005sleep}. In older adults, sleep disorders could be contributing to neuro-degenerative disorders \cite{naismith2011sleep}. 

Polysomnogram (PSG) is widely used as a diagnostic tool to determine sleep disorders. Polysomnography incorporates the recording and inspection of multiple physiological signals during sleep, like electroencephalogram (EEG), electro-oculogram (EOG) and electrocardiogram (ECG). The physiological variables and body movements help experts determine what stages of sleep the subjects have reached and their duration, depending on standard classification guidelines \cite{hobson1969manual}. 


Many studies in the past few decades have focused on developing automatic sleep stage classification (ASSC) frameworks. Multi-channel systems incorporating EEG, EOG and EMG \cite{cen2017deep,virkkala2007automatic}, single-channel systems with EEG \cite{vilamala2017deep} and multi-modal systems with smartphone and wearable sensors \cite{chen2017multimodal} have been explored. While state-of-the-art systems mostly employ feature extraction based methods along with machine learning algorithms \cite{boostani2017comparative}, recent papers have adopted the use of Deep Neural Networks (DNN) in an end-to-end fashion \cite{supratak2017deepsleepnet}. Digital filtering \cite{aboalayon2016entropy} and Wavelet Transform \cite{liu2017learning,sharma2018accurate}, along with its variants \cite{HASSAN2016}, have been common pre-processing steps utilized in previous studies. A number of these methods extract time \cite{aboalayon2016entropy}, frequency \cite{HASSAN2016} or time-frequency features from each decomposed band of the signal \cite{jiang2019robust}. These features are hand-engineered depending on the study population and therefore may not generalize well. Moreover, there is no consensus on which features are the most discriminatory for the sleep stage classification task \cite{boostani2017comparative}. Deep learning methods are known to automatically learn generalizing features, which have been leveraged in a number of different tasks in the past \cite{he2015delving}. Recent deep learning approaches for sleep staging have explored end-to-end systems using 1-dimensional Convolutional Neural Networks (1DCNN), Bi-directional Long Short Term Memory Units \cite{supratak2017deepsleepnet,phan2018automatic},
and also 2DCNNs with prior spectral decomposition to represent EEG as 2D images \cite{vilamala2017deep}. In \cite{manzano2017deep}, the EEG signal was first pre-processed with a notch and bandpass filter and followed by two separate CNN and Multi-layer Perceptron (MLP) architectures using time domain and frequency domain signals. Deep complex valued CNNs with C3-A2 EEG signals were employed in \cite{zhang2018automatic}. A 12-layer deep 1DCNN architecture for feature extraction was explored in \cite{sors2018convolutional}, while in \cite{supratak2017deepsleepnet} a 4-layer deep architecture was deployed.

In this paper, we introduce a 34-layer deep 1DCNN residual architecture for end-to-end classification. Residual architectures (ResNet) were proposed to resolve the vanishing gradient problem arising from the training of deeper CNN models \cite{he2015delving}. As the number of layers used in our model is significantly higher compared to previous studies on sleep staging, we utilize the ResNet architecture for our end-to-end system and compare its performance with state-of-the-art algorithms.
\vspace{-1mm}
\section{Dataset} \label{dataset}
A number of sleep health studies have been compiled into datasets that are used in quantitative experiments, i.e. the Sleep Heart Health Study (SHHS) \cite{sors2018convolutional}, Montreal Archive of Sleep Studies (MASS) \cite{supratak2017deepsleepnet} and the Physionet Sleep-EDF Database \cite{physionetPaper}. We use the Physionet Sleep-EDF Expanded Database in our experiments for its data volume and commonality in existing literature \cite{boostani2017comparative}. The database contains a total number of 61 polysomnography recordings from 42 subjects, sampled at 100 Hz. 30s data epochs from each recording were scored by sleep experts into one of the eight classes: \emph{AWA} (Awake), \emph{S1-S4} (intermediate stages), \emph{REM} (Rapid Eye Movement), \emph{MVT} (Movement Time) and 'Unscored' \cite{hobson1969manual}. The \emph{MVT} and unscored data epochs have been removed prior to our experiments since they don't correspond to any stages of sleep. We have also fused the \emph{S3} and \emph{S4} stages for 5-stage classification according to AASM \cite{berry2012aasm}.

The PhysioNet Sleep-EDF Expanded Database is assembled from two different studies. The first 39 recordings are from the Sleep Cassette (\emph{SC}) subset, which was collected from in-home sessions to study the effect of aging on health \cite{mourtazaev1995SC}. This portion of the dataset includes sleep data from 20 healthy subjects without any sleep-related medication. Each patient contributed two different recordings from two nights, except for patient number 13. The latter 22 data samples are from Sleep Telemetry (\emph{ST}) recording subset. It was collected during a hospital study on the effects of Temazepam on sleep. The data was acquired from 22 subjects for two different nights. Our experiments incorporate recordings from the nights without medication. The distribution of samples from various classes are shown in Table \ref{epochs}.

\begin{table}[t]
\centering
\caption{Epochs in each class for different subsets}
\label{epochs}
\resizebox{\linewidth}{!}{%
\begin{tabular}{@{}lcccccccc@{}}
\toprule
Subset & Subjects & S1 & S2 & S3 & S4 & REM & W & Total \\ \midrule
\multicolumn{1}{l|}{Sleep Cassette} & 20 & 2804 & 17799 & 3370 & 2333 & 7717 & 71887 & 105910 \\
\multicolumn{1}{l|}{Sleep Telemetry} & 22 & 2044 & 9493 & 1705 & 1440 & 4131 & 2285 & 21098 \\ \midrule
\multicolumn{1}{l|}{Sleep-EDF Total} & 42 & 4848 & 27292 & 5075 & 3773 & 11848 & 74172 & 127008 \\ \midrule
\multicolumn{1}{l|}{\emph{RS-task} - Train} &30 & 3499 & 18824 & 3339 & 2462 & 7857 & 46862 & 82843 \\
\multicolumn{1}{l|}{\emph{RS-task} - Test} &12 & 1349 & 8468 & 1736 & 1311 & 3991 & 27310 & 44165 \\ \bottomrule
\end{tabular}%
}
\end{table}

\section{Methodology} \label{method}
\subsection{Experimental Setup}
We design our experiments to address the robustness of deep learning models towards patient independent classification tasks. We have devised two test cases; the first is a multi-source task where a patient independent $71\%-29\%$ random split is performed on data from both \emph{SC} and \emph{ST}, from here on referred to as Random Split task (\emph{RS-task}). There are $42$ recordings ($25$ from $13$ \emph{SC} subjects \& $17$ from $17$ \emph{ST} subjects) in the training set and $19$ recordings ($14$ from $7$ \emph{SC} subjects \& $5$ from $5$ \emph{ST} subjects) in the test set of the \emph{RS-task}. The second case is when we train and test with the Sleep Cassette subset of the dataset with a $70\%-30\%$ random patient independent split, from here on referred to as Sleep Cassette task (\emph{SC-task}). This task has $27$ recordings from $14$ \emph{SC} subjects in the training set and $12$ recordings from $6$ \emph{SC} subjects in the test set. We perform each of these experiments with the FPz-Cz EEG channel data. The specific recording IDs and codes for each test case is available on \texttt{Github}
\footnote{Codes and supplementary material available at
\mdseries\urlstyle{tt}
\url{https://github.com/mHealthBuet/ASSC}}. Table \ref{epochs} contains the number of subjects and epochs from each class utilized in each of our experiments.


\begin{table*}[t]
\centering
\caption{Comparison of Proposed Method with Accuracy reported on Sleep-EDF Database (\emph{Sleep Cassette only})}
\label{compare}
\vspace{-2mm}
\resizebox{\textwidth}{!}{%
\begin{tabular}{@{}cccccc@{}}
\toprule
Approach &No. of Epochs & Split & Method & 5-Stage Acc & 6-Stage Acc \\ \midrule
K. Aboalayon 2016 \cite{aboalayon2016entropy} & 23806 Fpz-Cz 10s & 80\%-20\% Not independent & IIR-MMD-DT & \textbf{95.5\%} & \textbf{93.1\%} \\
N. Liu 2017 \cite{liu2017learning} & 60647 Fpz-Cz 30s & 90\%-10\% Not independent & DWT-CNN-MLP & - & 90.1\% \\
M. Sharma 2018 \cite{sharma2018accurate} & 85900 Single EEG 30s & 10 fold CV Not independent & Wavelet-SVM & 91.7\% & 91.5\% \\ \midrule
A. Vilamala 2017 \cite{vilamala2017deep} & Not Specified & 15-4 Independent & Spectral 2D-VGG16 & 86.0\% & - \\
A. Supratak 2017 \cite{supratak2017deepsleepnet} & 41950 Fpz-Cz 30s & 20 patients LOO-CV & CNN-BLSTM & 82.0\% & - \\
H. Phan 2018 \cite{phan2018automatic} & 46236 Fpz-Cz 30s & 20 patients LOO-CV & BLSTM-SVM & 82.5\% & - \\
D. Jiang 2019 \cite{jiang2019robust} & 54728 Fpz-Cz 30s & 20 patients LOO-CV & Multi.Decomp.-RF-HMM & 93.0\% & - \\
\textbf{Proposed Method} & 127008 Fpz-Cz 30s & 70\%-30\% Independent & Residual CNN & 91.4\% & 90.1\% \\ \bottomrule
\end{tabular}%
}
\end{table*}

\begin{table}[b]
\centering
\caption{Performance comparison with Abolayon et al. for $6$-stage classification}
\label{tableRes}
\vspace{-2mm}
\resizebox{\linewidth}{!}{%
\begin{tabular}{@{}ccccccc@{}}
\toprule
Method & \begin{tabular}[c]{@{}c@{}}Exp.\\ Task\end{tabular} &
\begin{tabular}[c]{@{}c@{}}S1\\ Sens.\end{tabular} & \begin{tabular}[c]{@{}c@{}}Avg\\ Sens.\end{tabular} & \begin{tabular}[c]{@{}c@{}}Avg\\ Spec.\end{tabular} & \begin{tabular}[c]{@{}c@{}}Epoch-Wise\\ Acc\end{tabular} & \begin{tabular}[c]{@{}c@{}}Patient-Wise\\ Acc\end{tabular} \\ \midrule
IIR-MMD-DT \cite{aboalayon2016entropy} & \emph{RS-task} & 15.6 & 58.2 & 95.6 & 83.8 & 80.3 \\
IIR-MMD-DT \cite{aboalayon2016entropy} & \emph{SC-task} & 26.4 & 61.9 & 94.8 & 84.3 & 84.3 \\ \midrule
Proposed Method & \emph{RS-task} & 43.8 & 66.5 & 97.9 & 88.8 & 85.5 \\
Proposed Method & \emph{SC-task} & 43.3 & 67.9 & 97.0 & 90.1 & 90.0 \\ \bottomrule
\end{tabular}%
}
\end{table}

\begin{figure}[t]
\includegraphics[width=60mm]{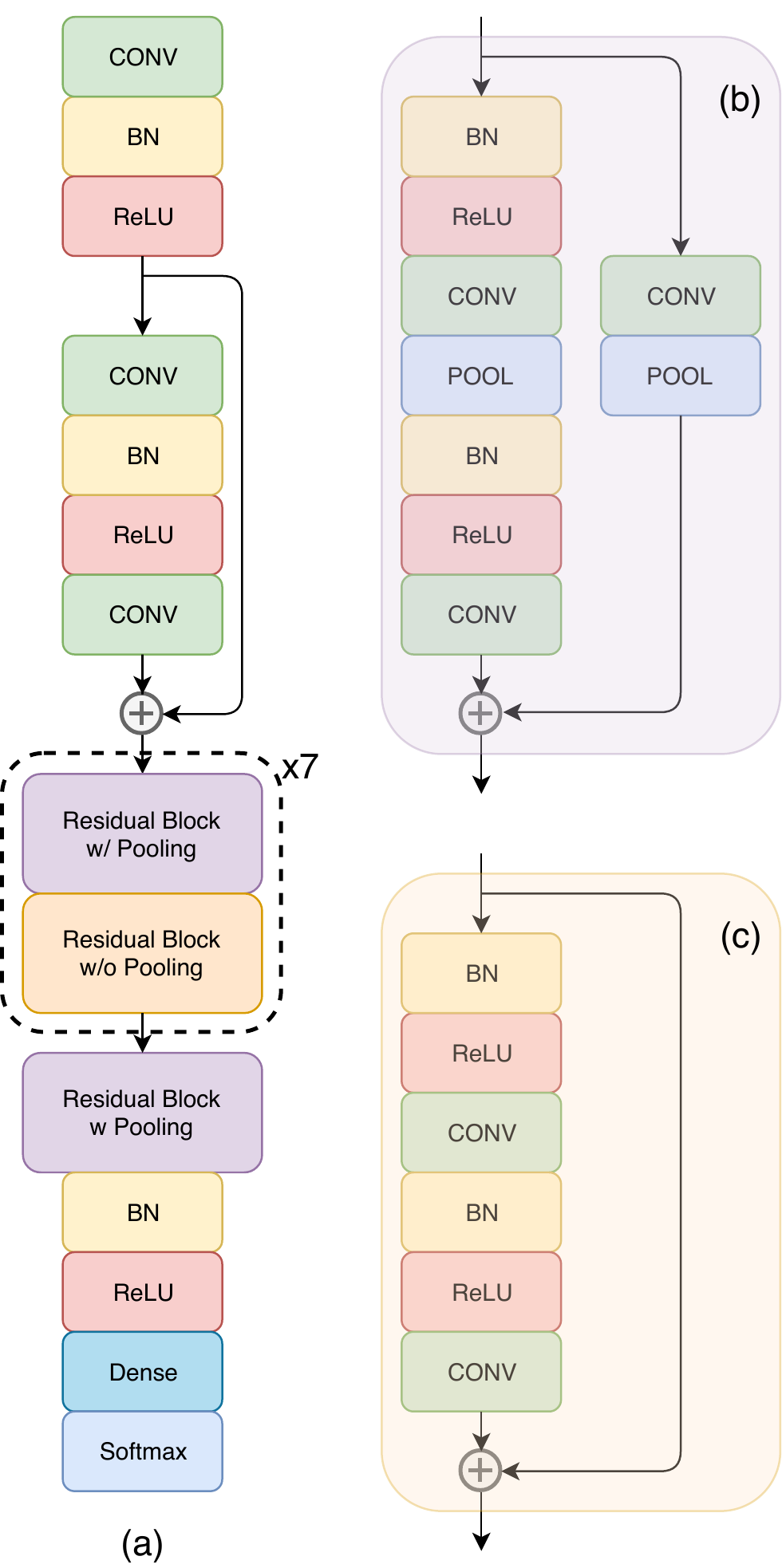}
\centering
\caption{The proposed residual architecture for $6$ class classification of sleep. (a) High-level overview of architecture with residual blocks. (b) Residual block with pooling layers both in the body and skip connection. (c) Residual block architecture without pooling layer.}
\label{ProposedModel}
\end{figure}

\subsection{Proposed Method}
Residual architectures have previously been employed for both image classification tasks \cite{he2016identity} and biomedical signal processing tasks \cite{rajpurkar2017cardiologist}. They were introduced to make optimization and convergence of very deep neural networks plausible. Training deeper neural networks poses challenges related to saturation and degradation of training accuracy with the increase of layers. 
Skip-connections in the residual blocks allow deeper CNN layers to approximate an identity operation instead of performing redundant transformations, by only passing the input tensor to the next block. We, therefore, employ skip-connections in our model for sequence-to-sequence sleep staging of each $30s$ EEG data epochs. Our proposed architecture has $34$ 1DCNN layers with the number of kernels in each CNN layer increasing with depth. Our model takes $30s$ unprocessed EEG time-domain signals as input and outputs hypnogram predictions for each instance. Fig. \ref{ProposedModel} portrays the high-level network architecture. The residual blocks with skip-connections follow a pre-activation design \cite{he2016identity} that performs addition operation between the input tensor and the transformed output prior to activation (Fig. \ref{ProposedModel}-c). This allows gradients to flow to shallower layers without aberration. Each CNN layer has a kernel size of $16$ and $64k$ filters where $k\in \{ {1,2,3,4}$\} increases by $1$ after every $8$ convolutional layers. Each pooling layer performs a max-pooling operation, sub-sampling the incoming data by $2$. The residual blocks with pooling have an extra CNN layer in the shortcut with a kernel size of $1$ and an equal number of filters as the second CNN layer in the corresponding block (Fig. \ref{ProposedModel}-b). It performs a linear combination operation to equalize the depth of the tensors before addition. Activations were dropped-out with a post-activation probability of $0.5$. The network was trained using Adam optimizer with a maximum learning rate of $0.001$ and we divide the learning rate by $10$ after every $10$ training epochs. Cross-entropy loss is weighted according to the dominance of classes in the training population to tackle class imbalance. During training, rolling shifts were performed to augment the training data for improved generalization.



\section{Results} \label{resDisc}
We compare the accuracy metric obtained by our proposed method on the \emph{SC-task} with that reported on recent papers in Table \ref{compare}. 
The evaluation methods used in literature can be divided into two types. The first is example splitting, where data epochs from the same patient are allowed to be in both the train and test sets; therefore the splits are \emph{not} patient independent \cite{aboalayon2016entropy,liu2017learning,sharma2018accurate}. The other way is to perform patient splitting where patients are split between training and testing to ensure a patient independent test set \cite{vilamala2017deep,supratak2017deepsleepnet,phan2018automatic}. Patient splitting can be performed at random \cite{vilamala2017deep} or in a leave-one-out cross-validation (LOO-CV) scheme where data from only one patient is used for evaluation \cite{jiang2019robust}. The number of epochs used in the experiments also require consideration. The performance achieved by our proposed method in the \emph{SC-task} is comparable with the accuracy acquired by state-of-the-art methods on the \emph{SC} subset of the Physionet Sleep-EDF database, surpassing most except for \cite{jiang2019robust} in patient independent tasks. For a fair comparison between patient independent and not independent splitting, we implement the method proposed by Aboalayon et al. \cite{aboalayon2016entropy}, which obtained the best accuracy reported for this problem. In our implementation, we used a Balanced Bagging ensemble of $71$ Decision Trees to compensate for class imbalance. Table \ref{tableRes} contains comparison of its acquired metrics with our proposed model. Our model exhibits a relative improvement of $6.8\%$ in epoch-wise accuracy on the \emph{SC-task} and $6.25\%$ on the \emph{RS-task}. We also evaluate how well our model is performing for individual patients; our model exhibits $6.76\%$ relative increase in patient-wise accuracy on the \emph{SC-task} and $6.45\%$ increase on the \emph{RS-task}.


\begin{figure}[b]
\includegraphics[width=0.9\linewidth, trim={0 12 0 10}, clip]{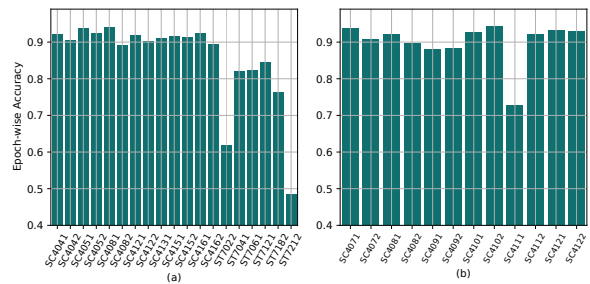}
\centering
\caption{Epoch-wise accuracy for different test set recordings in each task: (a) Random-split task (\emph{RS-task}) (b) Sleep Cassette task (\emph{SC-task}).}
\label{patacc}
\vspace{-5mm}
\end{figure}

\section{Discussion}
\subsection{Number of Epochs and Split}
The Sleep-EDF database, though widely employed in quantitative experiments, lacks a standard test bench for fair comparison between different ASSC frameworks. Evaluation metrics are susceptible to changes in class distributions, number of epochs used in the evaluation task and also the epoch selection scheme \cite{sors2018convolutional}. We can see a $8.8\%$ reduction in the accuracy of \cite{aboalayon2016entropy} in the $6$-stage \emph{SC-task} compared to the paper reported metric, which can be due to an increased number of data epochs and patient independent splitting during evaluation. Our \emph{SC-task} uses data from multiple patients (Fig \ref{patacc}) during test, which is more challenging compared to LOO-CV. This explains the accuracy difference between Jiang et al. \cite{jiang2019robust} and our proposed method (Table \ref{compare}).
\subsection{Difference between \emph{SC} \& \emph{ST} subsets}
\begin{figure}[t]
\includegraphics[width=0.82\linewidth,trim={0 10mm 0 0}]{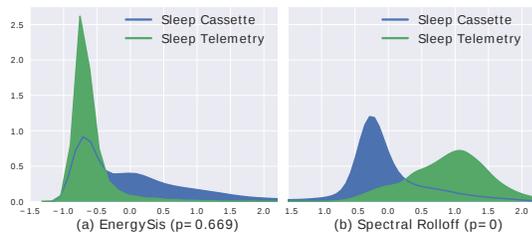}
\centering
\caption{Kernel Density Plots of two different features for Sleep Cassette and Sleep Telemetry subsets a) Scaled $\gamma$-band \emph{EnergySis} feature b) Scaled $\gamma$-band Spectral Rolloff feature.}
\label{distribution}
\vspace{-3mm}
\end{figure}

From Fig. \ref{patacc}-a we observe that our model exhibits a drop in accuracy for recordings which are from the \emph{ST} subset compared to the \emph{SC} subset. No such trend is visible when the model is trained and evaluated on the \emph{SC} subset (Fig \ref{patacc}-b). This can be attributed to the presence of data from two different sources in the \emph{RS-task} which was also discussed in \cite{sors2018convolutional}. To explore whether the \emph{SC} and \emph{ST} subsets in the Sleep-EDF database exhibit heterogeneous properties or not, we perform one-way ANOVA on two time-domain features (MMD \& \emph{EnergySis} \cite{aboalayon2016entropy}) and two spectral features (Spectral Rolloff \& Spectral Spread) extracted from $5$ frequency bands ($\delta,\theta,\alpha,\beta,\gamma$) of the Fpz-Cz channel.
Except for the $\gamma$-band \emph{EnergySis} feature (Fig. \ref{distribution}-a), all of the features exhibit a significant difference in the distribution with a $p \ll 0.001$. Fig \ref{distribution}-b shows significant difference between the density functions of $\gamma$-band Spectral Rolloff between \emph{SC} and \emph{ST}. This strengthens our argument regarding heterogeneity between the subsets \& explains the accuracy difference between \emph{RS-task} \& \emph{SC-task}. 

\section{Conclusion}
In this paper, we have proposed a residual architecture based ASSC framework. Our evaluation task comprised of an increased number of data epochs and subjects from multiple sources, compared to previously reported works. Experimental results showed that our model is robust towards noise perturbations which demonstrates relative improvement in epoch-wise average accuracy of $6.8\%$ on the \emph{SC-task} and $6.3\%$ on the \emph{RS-task} compared to Aboalayon et al. \cite{aboalayon2016entropy}.

\section*{Acknowledgment}
We would like to thank the Department of EEE \& BME, BUET and Brain Station 23 (Dhaka, Bangladesh) for supporting this research. The TITAN Xp GPU used for this work was donated by the NVIDIA Corporation.

\bibliographystyle{IEEEtran}
\bibliography{refs.bib}

\section*{Appendix}
\begin{table}[H]
\centering
\caption{Confusion Matrix of our proposed method for 5 Class Classification on \emph{SC-task}}
\label{confMatSC}
\resizebox{.8\linewidth}{!}{%
\begin{tabular}{cccccc}
 & S1 & S2 & S3 & REM & AWA \\ \cline{2-6} 
\multicolumn{1}{r|}{S1} & \textbf{44.23} & 16.27 & 0.66 & 32.55 & 6.3 \\
\multicolumn{1}{r|}{S2} & 4.48 & \textbf{79.72} & 5.33 & 4.12 & 6.35 \\
\multicolumn{1}{r|}{S3} & 0.11 & 21.25 & \textbf{76.58} & 0 & 2.06 \\
\multicolumn{1}{r|}{REM} & 7.05 & 8.53 & 0 & \textbf{75.97} & 8.45 \\
\multicolumn{1}{r|}{AWA} & 2.44 & 0.15 & 0.03 & 1 & \textbf{96.38}
\end{tabular}%
}
\vspace{5.5mm}
\centering
\caption{Implementation Details of Proposed Architecture}
\resizebox{.93\linewidth}{!}{%
\begin{tabular}{@{}lccccr@{}}
\toprule
Layer Name & Layer Type & Width & Channels & Params & Input Layer Names \\ \midrule
input\_1 & InputLayer & 3000 & 1 & 0 &  \\
conv1d\_1 & Conv1D & 3000 & 64 & 1088 & input\_1 \\
batch\_normalization\_1 & BatchNorm & 3000 & 64 & 256 & conv1d\_1 \\
scale\_1 & Scale & 3000 & 64 & 128 & batch\_normalization\_1 \\
activation\_1 & Activation & 3000 & 64 & 0 & scale\_1 \\
conv1d\_2 & Conv1D & 3000 & 64 & 65536 & activation\_1 \\
batch\_normalization\_2 & BatchNorm & 3000 & 64 & 256 & conv1d\_2 \\
scale\_2 & Scale & 3000 & 64 & 128 & batch\_normalization\_2 \\
activation\_2 & Activation & 3000 & 64 & 0 & scale\_2 \\
dropout\_1 & Dropout & 3000 & 64 & 0 & activation\_2 \\
conv1d\_3 & Conv1D & 3000 & 64 & 65536 & dropout\_1 \\
add\_1 & Add & 3000 & 64 & 0 & conv1d\_3,activation\_1 \\
batch\_normalization\_3 & BatchNorm & 3000 & 64 & 256 & add\_1 \\
scale\_3 & Scale & 3000 & 64 & 128 & batch\_normalization\_3 \\
activation\_3 & Activation & 3000 & 64 & 0 & scale\_3 \\
dropout\_2 & Dropout & 3000 & 64 & 0 & activation\_3 \\
conv1d\_4 & Conv1D & 3000 & 64 & 65536 & dropout\_2 \\
max\_pooling1d\_1 & MaxPooling1D & 1500 & 64 & 0 & conv1d\_4 \\
batch\_normalization\_4 & BatchNorm & 1500 & 64 & 256 & max\_pooling1d\_1 \\
scale\_4 & Scale & 1500 & 64 & 128 & batch\_normalization\_4 \\
activation\_4 & Activation & 1500 & 64 & 0 & scale\_4 \\
dropout\_3 & Dropout & 1500 & 64 & 0 & activation\_4 \\
conv1d\_6 & Conv1D & 3000 & 64 & 4096 & add\_1 \\
conv1d\_5 & Conv1D & 1500 & 64 & 65536 & dropout\_3 \\
max\_pooling1d\_2 & MaxPooling1D & 1500 & 64 & 0 & conv1d\_6 \\
add\_2 & Add & 1500 & 64 & 0 & conv1d\_5.max\_pooling1d\_2 \\
batch\_normalization\_5 & BatchNorm & 1500 & 64 & 256 & add\_2 \\
scale\_5 & Scale & 1500 & 64 & 128 & batch\_normalization\_5 \\
activation\_5 & Activation & 1500 & 64 & 0 & scale\_5 \\
dropout\_4 & Dropout & 1500 & 64 & 0 & activation\_5 \\
conv1d\_7 & Conv1D & 1500 & 64 & 65536 & dropout\_4 \\
batch\_normalization\_6 & BatchNorm & 1500 & 64 & 256 & conv1d\_7 \\
scale\_6 & Scale & 1500 & 64 & 128 & batch\_normalization\_6 \\
activation\_6 & Activation & 1500 & 64 & 0 & scale\_6 \\
dropout\_5 & Dropout & 1500 & 64 & 0 & activation\_6 \\
conv1d\_8 & Conv1D & 1500 & 64 & 65536 & dropout\_5 \\
add\_3 & Add & 1500 & 64 & 0 & conv1d\_8,add\_2 \\
batch\_normalization\_7 & BatchNorm & 1500 & 64 & 256 & add\_3 \\
scale\_7 & Scale & 1500 & 64 & 128 & batch\_normalization\_7 \\
activation\_7 & Activation & 1500 & 64 & 0 & scale\_7 \\
dropout\_6 & Dropout & 1500 & 64 & 0 & activation\_7 \\
conv1d\_9 & Conv1D & 1500 & 64 & 65536 & dropout\_6 \\
max\_pooling1d\_3 & MaxPooling1D & 750 & 64 & 0 & conv1d\_9 \\
batch\_normalization\_8 & BatchNorm & 750 & 64 & 256 & max\_pooling1d\_3 \\
scale\_8 & Scale & 750 & 64 & 128 & batch\_normalization\_8 \\
activation\_8 & Activation & 750 & 64 & 0 & scale\_8 \\
dropout\_7 & Dropout & 750 & 64 & 0 & activation\_8 \\
conv1d\_11 & Conv1D & 1500 & 128 & 8192 & add\_3 \\
conv1d\_10 & Conv1D & 750 & 128 & 131072 & dropout\_7 \\
max\_pooling1d\_4 & MaxPooling1D & 750 & 128 & 0 & conv1d\_11 \\
add\_4 & Add & 750 & 128 & 0 & conv1d\_10,max\_pooling1d\_4 \\ 
batch\_normalization\_9 & BatchNorm & 750 & 128 & 512 & add\_4 \\
scale\_9 & Scale & 750 & 128 & 256 & batch\_normalization\_9 \\
activation\_9 & Activation & 750 & 128 & 0 & scale\_9 \\
dropout\_8 & Dropout & 750 & 128 & 0 & activation\_9 \\
conv1d\_12 & Conv1D & 750 & 128 & 262144 & dropout\_8 \\
batch\_normalization\_10 & BatchNo & 750 & 128 & 512 & conv1d\_12 \\
scale\_10 & Scale & 750 & 128 & 256 & batch\_normalization\_10 \\
activation\_10 & Activation & 750 & 128 & 0 & scale\_10 \\
dropout\_9 & Dropout & 750 & 128 & 0 & activation\_10 \\
conv1d\_13 & Conv1D & 750 & 128 & 262144 & dropout\_9 \\
add\_5 & Add & 750 & 128 & 0 & conv1d\_13,add\_4 \\
batch\_normalization\_11 & BatchNo & 750 & 128 & 512 & add\_5 \\
scale\_11 & Scale & 750 & 128 & 256 & batch\_normalization\_11 \\
activation\_11 & Activation & 750 & 128 & 0 & scale\_11 \\
dropout\_10 & Dropout & 750 & 128 & 0 & activation\_11 \\
conv1d\_14 & Conv1D & 750 & 128 & 262144 & dropout\_10 \\
max\_pooling1d\_5 & MaxPooling1D & 375 & 128 & 0 & conv1d\_14 \\
batch\_normalization\_12 & BatchNo & 375 & 128 & 512 & max\_pooling1d\_5 \\
scale\_12 & Scale & 375 & 128 & 256 & batch\_normalization\_12 \\
activation\_12 & Activation & 375 & 128 & 0 & scale\_12 \\
dropout\_11 & Dropout & 375 & 128 & 0 & activation\_12 \\
conv1d\_16 & Conv1D & 750 & 128 & 16384 & add\_5 \\
conv1d\_15 & Conv1D & 375 & 128 & 262144 & dropout\_11 \\
max\_pooling1d\_6 & MaxPooling1D & 375 & 128 & 0 & conv1d\_16 \\
add\_6 & Add & 375 & 128 & 0 & conv1d\_15,max\_pooling1d\_6 \\
batch\_normalization\_13 & BatchNo & 375 & 128 & 512 & add\_6 \\
scale\_13 & Scale & 375 & 128 & 256 & batch\_normalization\_13 \\
activation\_13 & Activation & 375 & 128 & 0 & scale\_13 \\
dropout\_12 & Dropout & 375 & 128 & 0 & activation\_13 \\
conv1d\_17 & Conv1D & 375 & 128 & 262144 & dropout\_12 \\
batch\_normalization\_14 & BatchNo & 375 & 128 & 512 & conv1d\_17 \\
scale\_14 & Scale & 375 & 128 & 256 & batch\_normalization\_14 \\
activation\_14 & Activation & 375 & 128 & 0 & scale\_14 \\
dropout\_13 & Dropout & 375 & 128 & 0 & activation\_14 \\
conv1d\_18 & Conv1D & 375 & 128 & 262144 & dropout\_13 \\
add\_7 & Add & 375 & 128 & 0 & conv1d\_18,add\_6 \\
batch\_normalization\_15 & BatchNo & 375 & 128 & 512 & add\_7 \\
scale\_15 & Scale & 375 & 128 & 256 & batch\_normalization\_15 \\
activation\_15 & Activation & 375 & 128 & 0 & scale\_15 \\ \bottomrule
\end{tabular}%
}
\end{table}

\begin{table}[t]
\centering
\caption{Implementation Details of Proposed Architecture (\emph{Continued})}
\resizebox{.94\linewidth}{!}{%
\begin{tabular}{lccccr}
\toprule
Layer Name & Layer Type & Width & Channels & Params & Input Layer Names \\ \midrule
dropout\_14 & Dropout & 375 & 128 & 0 & activation\_15 \\
conv1d\_19 & Conv1D & 375 & 128 & 262144 & dropout\_14 \\
max\_pooling1d\_7 & MaxPooling1D & 187 & 128 & 0 & conv1d\_19 \\
batch\_normalization\_16 & BatchNo & 187 & 128 & 512 & max\_pooling1d\_7 \\
scale\_16 & Scale & 187 & 128 & 256 & batch\_normalization\_16 \\
activation\_16 & Activation & 187 & 128 & 0 & scale\_16 \\
dropout\_15 & Dropout & 187 & 128 & 0 & activation\_16 \\
conv1d\_21 & Conv1D & 375 & 192 & 24576 & add\_7 \\
conv1d\_20 & Conv1D & 187 & 192 & 393216 & dropout\_15 \\
max\_pooling1d\_8 & MaxPooling1D & 187 & 192 & 0 & conv1d\_21 \\
add\_8 & Add & 187 & 192 & 0 & conv1d\_20,max\_pooling1d\_8 \\ 
batch\_normalization\_17 & BatchNo & 187 & 192 & 768 & add\_8 \\
scale\_17 & Scale & 187 & 192 & 384 & batch\_normalization\_17 \\
activation\_17 & Activation & 187 & 192 & 0 & scale\_17 \\
dropout\_16 & Dropout & 187 & 192 & 0 & activation\_17 \\
conv1d\_22 & Conv1D & 187 & 192 & 589824 & dropout\_16 \\
batch\_normalization\_18 & BatchNo & 187 & 192 & 768 & conv1d\_22 \\
scale\_18 & Scale & 187 & 192 & 384 & batch\_normalization\_18 \\
activation\_18 & Activation & 187 & 192 & 0 & scale\_18 \\
dropout\_17 & Dropout & 187 & 192 & 0 & activation\_18 \\
conv1d\_23 & Conv1D & 187 & 192 & 589824 & dropout\_17 \\
add\_9 & Add & 187 & 192 & 0 & conv1d\_23,add\_8 \\
batch\_normalization\_19 & BatchNo & 187 & 192 & 768 & add\_9 \\
scale\_19 & Scale & 187 & 192 & 384 & batch\_normalization\_19 \\
activation\_19 & Activation & 187 & 192 & 0 & scale\_19 \\
dropout\_18 & Dropout & 187 & 192 & 0 & activation\_19 \\
conv1d\_24 & Conv1D & 187 & 192 & 589824 & dropout\_18 \\
max\_pooling1d\_9 & MaxPooling1D & 93 & 192 & 0 & conv1d\_24 \\
batch\_normalization\_20 & BatchNo & 93 & 192 & 768 & max\_pooling1d\_9 \\
scale\_20 & Scale & 93 & 192 & 384 & batch\_normalization\_20 \\
activation\_20 & Activation & 93 & 192 & 0 & scale\_20 \\
dropout\_19 & Dropout & 93 & 192 & 0 & activation\_20 \\
conv1d\_26 & Conv1D & 187 & 192 & 36864 & add\_9 \\
conv1d\_25 & Conv1D & 93 & 192 & 589824 & dropout\_19 \\
max\_pooling1d\_10 & MaxPooling1D & 93 & 192 & 0 & conv1d\_26 \\
add\_10 & Add & 93 & 192 & 0 & conv1d\_25,max\_pooling1d\_10 \\
batch\_normalization\_21 & BatchNo & 93 & 192 & 768 & add\_10 \\
scale\_21 & Scale & 93 & 192 & 384 & batch\_normalization\_21 \\
activation\_21 & Activation & 93 & 192 & 0 & scale\_21 \\
dropout\_20 & Dropout & 93 & 192 & 0 & activation\_21 \\
conv1d\_27 & Conv1D & 93 & 192 & 589824 & dropout\_20 \\
batch\_normalization\_22 & BatchNo & 93 & 192 & 768 & conv1d\_27 \\
scale\_22 & Scale & 93 & 192 & 384 & batch\_normalization\_22 \\
activation\_22 & Activation & 93 & 192 & 0 & scale\_22 \\
dropout\_21 & Dropout & 93 & 192 & 0 & activation\_22 \\
conv1d\_28 & Conv1D & 93 & 192 & 589824 & dropout\_21 \\
add\_11 & Add & 93 & 192 & 0 & conv1d\_28,add\_10 \\
batch\_normalization\_23 & BatchNo & 93 & 192 & 768 & add\_11 \\
scale\_23 & Scale & 93 & 192 & 384 & batch\_normalization\_23 \\
activation\_23 & Activation & 93 & 192 & 0 & scale\_23 \\
dropout\_22 & Dropout & 93 & 192 & 0 & activation\_23 \\
conv1d\_29 & Conv1D & 93 & 192 & 589824 & dropout\_22 \\
max\_pooling1d\_11 & MaxPooling1D & 46 & 192 & 0 & conv1d\_29 \\
batch\_normalization\_24 & BatchNo & 46 & 192 & 768 & max\_pooling1d\_11 \\
scale\_24 & Scale & 46 & 192 & 384 & batch\_normalization\_24 \\
activation\_24 & Activation & 46 & 192 & 0 & scale\_24 \\
dropout\_23 & Dropout & 46 & 192 & 0 & activation\_24 \\
conv1d\_31 & Conv1D & 93 & 256 & 49152 & add\_11 \\
conv1d\_30 & Conv1D & 46 & 256 & 786432 & dropout\_23 \\
max\_pooling1d\_12 & MaxPooling1D & 46 & 256 & 0 & conv1d\_31 \\
add\_12 & Add & 46 & 256 & 0 & conv1d\_30,max\_pooling1d\_12 \\
batch\_normalization\_25 & BatchNo & 46 & 256 & 1024 & add\_12 \\
scale\_25 & Scale & 46 & 256 & 512 & batch\_normalization\_25 \\
activation\_25 & Activation & 46 & 256 & 0 & scale\_25 \\
dropout\_24 & Dropout & 46 & 256 & 0 & activation\_25 \\
conv1d\_32 & Conv1D & 46 & 256 & 1048576 & dropout\_24 \\
batch\_normalization\_26 & BatchNo & 46 & 256 & 1024 & conv1d\_32 \\
scale\_26 & Scale & 46 & 256 & 512 & batch\_normalization\_26 \\
activation\_26 & Activation & 46 & 256 & 0 & scale\_26 \\
dropout\_25 & Dropout & 46 & 256 & 0 & activation\_26 \\
conv1d\_33 & Conv1D & 46 & 256 & 1048576 & dropout\_25 \\
add\_13 & Add & 46 & 256 & 0 & conv1d\_33,add\_12 \\
batch\_normalization\_27 & BatchNo & 46 & 256 & 1024 & add\_13 \\
scale\_27 & Scale & 46 & 256 & 512 & batch\_normalization\_27 \\
activation\_27 & Activation & 46 & 256 & 0 & scale\_27 \\
dropout\_26 & Dropout & 46 & 256 & 0 & activation\_27 \\
conv1d\_34 & Conv1D & 46 & 256 & 1048576 & dropout\_26 \\
max\_pooling1d\_13 & MaxPooling1D & 23 & 256 & 0 & conv1d\_34 \\
batch\_normalization\_28 & BatchNo & 23 & 256 & 1024 & max\_pooling1d\_13 \\
scale\_28 & Scale & 23 & 256 & 512 & batch\_normalization\_28 \\
activation\_28 & Activation & 23 & 256 & 0 & scale\_28 \\
dropout\_27 & Dropout & 23 & 256 & 0 & activation\_28 \\
conv1d\_36 & Conv1D & 46 & 256 & 65536 & add\_13 \\
conv1d\_35 & Conv1D & 23 & 256 & 1048576 & dropout\_27 \\
max\_pooling1d\_14 & MaxPooling1D & 23 & 256 & 0 & conv1d\_36 \\
add\_14 & Add & 23 & 256 & 0 & conv1d\_35,max\_pooling1d\_14 \\
batch\_normalization\_29 & BatchNo & 23 & 256 & 1024 & add\_14 \\
scale\_29 & Scale & 23 & 256 & 512 & batch\_normalization\_29 \\
activation\_29 & Activation & 23 & 256 & 0 & scale\_29 \\
dropout\_28 & Dropout & 23 & 256 & 0 & activation\_29 \\
conv1d\_37 & Conv1D & 23 & 256 & 1048576 & dropout\_28 \\
batch\_normalization\_30 & BatchNo & 23 & 256 & 1024 & conv1d\_37 \\
scale\_30 & Scale & 23 & 256 & 512 & batch\_normalization\_30 \\
activation\_30 & Activation & 23 & 256 & 0 & scale\_30 \\
dropout\_29 & Dropout & 23 & 256 & 0 & activation\_30 \\
conv1d\_38 & Conv1D & 23 & 256 & 1048576 & dropout\_29 \\
add\_15 & Add & 23 & 256 & 0 & conv1d\_38,add\_14 \\ 
batch\_normalization\_31 & BatchNo & 23 & 256 & 1024 & add\_15 \\
scale\_31 & Scale & 23 & 256 & 512 & batch\_normalization\_31 \\
activation\_31 & Activation & 23 & 256 & 0 & scale\_31 \\
dropout\_30 & Dropout & 23 & 256 & 0 & activation\_31 \\
conv1d\_39 & Conv1D & 23 & 256 & 1048576 & dropout\_30 \\
max\_pooling1d\_15 & MaxPooling1D & 11 & 256 & 0 & conv1d\_39 \\
batch\_normalization\_32 & BatchNo & 11 & 256 & 1024 & max\_pooling1d\_15 \\
scale\_32 & Scale & 11 & 256 & 512 & batch\_normalization\_32 \\
activation\_32 & Activation & 11 & 256 & 0 & scale\_32 \\
dropout\_31 & Dropout & 11 & 256 & 0 & activation\_32 \\
conv1d\_41 & Conv1D & 23 & 512 & 131072 & add\_15 \\
conv1d\_40 & Conv1D & 11 & 512 & 2097152 & dropout\_31 \\
max\_pooling1d\_16 & MaxPooling1D & 11 & 512 & 0 & conv1d\_41 \\
add\_16 & Add & 11 & 512 & 0 & conv1d\_40,max\_pooling1d\_16 \\
batch\_normalization\_33 & BatchNo & 11 & 512 & 2048 & add\_16 \\
scale\_33 & Scale & 11 & 512 & 1024 & batch\_normalization\_33 \\
activation\_33 & Activation & 11 & 512 & 0 & scale\_33 \\
flatten\_1 & Flatten & 5632 & 0 & 0 & activation\_33 \\
dense\_1 & Dense & 5 & 0 & 28165 & flatten\_1 \\ \bottomrule
\end{tabular}%
}
\end{table}

\end{document}